\documentclass[runningheads,a4paper]{llncs}

\usepackage{amssymb}
\usepackage{amsmath}
\usepackage{graphicx}
\usepackage{enumitem}
\usepackage{stfloats} 
\usepackage{url} 
\urldef{\mailsa}\path|{n.alaya, m.lamolle}@iut.univ-paris8.fr|    
\urldef{\mailsb}\path|sadok.benyahia@fst.rnu.tn|
\newcommand{\keywords}[1]{\par\addvspace\baselineskip
\noindent\keywordname\enspace\ignorespaces#1}

\begin{document}

\mainmatter  

\title{Towards Unveiling the Ontology Key Features Altering Reasoner Performances}

\titlerunning{Ontology Key Features Altering Reasoner Performances}

%
%
\author{Nourh\`{e}ne Alaya
\and Sadok Ben Yahia \and Myriam Lamolle}
\authorrunning{Lecture Notes in Computer Science: Authors' Instructions}
%
\institute{LIPAH, Faculty of Sciences of Tunis, University of Tunis EL Manar, Tunisia.\\
LIASD, IUT of Montreuil, University of Paris 8, France.\\
\mailsa , 
\mailsb \\
}

%
%

\toctitle{Lecture Notes in Computer Science}
\tocauthor{Authors' Instructions}
\maketitle

\begin{abstract}
Reasoning with ontologies is one of the core fields of research in Description Logics. A variety of efficient reasoner with highly optimized algorithms have been developed to allow  inference tasks on expressive ontology languages such as OWL(DL). However, reasoner reported computing times have exceeded and sometimes fall behind the expected theoretical values. From an empirical perspective, it is not yet well understood, which particular aspects in the ontology are reasoner performance degrading factors. In this paper, we conducted an investigation  about state of art works that attempted to portray potential correlation between reasoner empirical behaviour and particular ontological features. These works were analysed and then broken down into categories. Further, we proposed a set of ontology features covering a broad range of structural and syntactic ontology characteristics. We claim that these features are good indicators of the ontology hardness level against reasoning tasks. 
\end{abstract}

\section{Introduction}
Ontologies are used as conceptual models, for data integration, or to directly represent information in a variety of domain areas. Considered as the centerpiece of knowledge description in the semantic web, ontologies continue to gain in importance as well as in size and complexity. The proliferation of ontologies poses  new  compelling challenges for the semantic web applications. The high expressivity of the ontology languages, such as OWL and in particular OWL 2, increased the computational complexity of inference tasks. For instance, it has been shown that the complexity of the consistency checking of $\mathcal{SROIQ}$ ontologies, the description logic (DL) underlying OWL 2, is of worst-case 2NExpTime-complete \cite{Horrocks06}. Therefore, a considerable efforts has been devoted to make reasoning feasible in practice. A number of highly-optimized reasoners have been developed \cite{Hermit12,Pellet07,Tsarkov06}, that support reasoning about ontologies written in expressive description logics.\\

However, empirical studies have revealed the unpredictable nature  of reasoner's behaviours when dealing with individual ontologies. In several cases, the theoretical worst case complexity does not necessarily unveil real-world performances. Reported computing times could exceed or eventually fall behind expected values \cite{ORe2013Results}. In one hand, Mart\"{i}n-Recuerda et al. \cite{Towards11} highlighted that reasoning in practice is far less complex then the established theoretical complexity. Interestingly enough, even with fairly expressive fragments of OWL 2, acceptable reasoning performances could be achieved. Yet, they admitted that it is still not well understood why reasoning is feasible in practice. On the other hand, Gon\c{c}alves et al. \cite{Hotspots12} outlined the performance variability phenomena with OWL ontologies. They reported three particular situations that a user could face when attempting to reason about an ontology: (\textit{i}) For one test case ontology, switching the reasoner can degrade reasoning time from seconds to none termination; (\textit{ii}) Ontologies with the same size and expressivity would spend wildly different ranges of computational time on the same reasoner; (\textit{iii}) An insignificant change to an OWL ontology, would increase or probably decrease reasoning time on one reasoner.\\

As far as one of these situations happens, often no feedback is returned back to the user. Commonly, the latter one will keep shifting reasoners until finding the suitable one. Others would try to adjust their ontology hopping for an improvement, but running the risk of making the matters worse. Throughout, it seems that all of these attempts could be time and effort consuming and may not lead to significant answers. In fact, as previously highlighted in \cite{WangP07}, the actual problem is that both ontology novice and expert users are lacking of theory and tool support helping analysing reasoner's behaviours against case study ontologies. Obviously, a better understanding of ontology complexity factors that may  trigger difficulties to reasoners is of a compelling need. In addition, pointing out what makes reasoning hard in practice, can guide the modelling process of ontologies as to avoid the reasoning performance degrading factors. Moreover, existing ontologies may be revised towards efficient reasoning by detecting, and even repairing its critical components.\\

In this paper, we carried out an investigation about existing methods and tools that intended to identify potential correlation between reasoner empirical behaviour and particular ontological features. We tried to give users an overall view of the state of art in this field. The pioneering works were analysed and then brook down into categories. Our investigations have lead us to propose a set of ontology features covering a broad rang of structural and syntactic ontology characteristics. We claim that these features would be key indicators of the ontology hardness level against reasoning tasks. \\

The rest of the paper is organized as follows: Section \ref{background} briefly describes basic terms such as ontology, description logic and reasoning. Section \ref{works} scrutinizes the related work approaches. Section \ref{ontofeatures} details our proposed features.Our concluding remarks as well as a sketching of future works are given in Section \ref{conclusion}.

\section{Background}\label{background}
The term \emph{expressivity} could be confusing in the ontology field. In fact, it has different meaning depending on the context of its use \cite{theevaluation07}. For instance, expressivity in the field of ontology quality assessment is used to describe the ontology knowledge richness \texttt{w.r.t.}  the domain under conceptualization. However, expressivity at the knowledge representation (KR) field is often used to characterize the KR language grammar, available for the ontology authoring. In this paper, we focus on the ontology expressivity at the KR language level as we are aiming at identified domain independent features, likely to outline the hardness of the ontology for the reasoning tasks.\\

In the remainder, we first recall the basics of OWL language and the Description Logics and then, we remind the main concepts of reasoning with ontologies.
\subsection{OWL Ontologies and Description Logics}
In this paper, we focus on OWL ontologies. OWL is one of the most widely used ontology languages, it has excellent tool support in terms of editors and reasoners. The latest version of OWL is OWL 2, which became a W3C recommendation in October 2009 and is based on the highly expressive Description Logics $\mathcal{SROIQ}$ \cite{Horrocks06}.  This logical gives statements made in OWL a precisely defined meaning and, for a given ontology, makes it possible to use automated reasoning to compute whether or not a statement follows from the ontology and various other reasoning tasks that we will explain latter in this section. In the remainder of this paper, by OWL we will mean OWL 2.\\

Generally speaking, Description  Logics  (DLs) are  subsets  of  the  first-order predication logic (FOL) \cite{DLBook03}. A DL ontology $\mathcal{O}$ is composed of a set of asserted axioms, analogous to FOL formulae, describing relationships between terms in a domain of interest. These terms are basically concept, role, and individual names, organised respectively in three sets $N_{C}$ , $N_{R}$, and $N_{I}$. The union of these sets, that is, the set of all terms mentioned in $\mathcal{O}$, is called the signature of $\mathcal{O}$, and denoted $\widetilde{\mathcal{O}}$.
In DL, an ontology is basically defined as a knowledge base $\mathcal{K =\langle T, R, A\rangle}$; where $\mathcal{T}$ denotes \emph{TBox}, which comprises \emph{terminological} axioms describing concepts, $\mathcal{R}$ denotes \emph{RBox} for axioms describing roles; and $\mathcal{A}$ stands for \emph{ABox}, which is the assertional part of knowledge base describing individuals. 
\begin{table}
    \begin{center}
   \caption{Basic OWL axioms from TBox, RBox and ABox, by referring to []}
  \begin{tabular}{llll}
  \hline 
  \textbf{Category} & \textbf{OWL Axiom} & \textbf{DL Syntax} & \textbf{Example} \\ 
  \hline 
\emph{TBox}& subClassOf & $A \equiv C$ & Man $\equiv$ Human $\sqcap$ Male  \\
           & equivalentClass & $ C \sqsubseteq D$& Human $\sqsubseteq$ Animal $\sqcap$ Biped \\
           & disjointWith & $ C \equiv \neg D$  & $ Male \equiv \neg Female$ \\
\emph{RBox} & equivalentProperty & $R \equiv S$ & cost $\equiv$ price \\
            & subPropertyOf & $R \sqsubseteq S$ & hasDaughter `$\sqsubseteq$ hasChild \\
            & inverseOf & $R \equiv S^{-}$ & hasChild $\equiv$ hasParent$^{-}$ \\
            & transitiveProperty & $R^{+} \sqsubseteq R $ &ancetor$^{+} \sqsubseteq$ ancetor \\
            & functionalProperty & $\top \sqsubseteq$ $\leq1R $&  $\top \sqsubseteq$ $\leq1$ hasMother \\
            & inverseFunctionalProperty & $\top \sqsubseteq$ $\leq1R^{-}$& $\top \sqsubseteq$ $\leq1$ hasSSN$^{-}$ \\
\emph{ABox} & Concept assertion & $C(a)$& Human({Peter}) \\
             & Role assertion & $R(a,b)$ & hasMother({Peter}, {Mary})\\
             & sameIndividualAs &  $ \{ x_{1} \} \equiv \{ x_{2} \}$ & {President-Kennedy} $\equiv $ {J-F-K} \\
             & differentFrom &  $\{x_1\} \equiv \neg \{x_2\}$& {Peter} $ \equiv \neg$ {John}\\
  \hline 
  \end{tabular} 
    \label{tab:TableAxiom}
  \end{center}
\end{table}
Table~\ref{tab:TableAxiom} shows some types of OWL axioms belonging to different categories, where $A$ and $B$ can be named concepts (also called atomic concepts), $C$ and $D$ are complex concept descriptions \cite{DLBook03}, $R$ and $S$ are role names or descriptions and $a$, $b$ and $x$ are names for individuals. Complex concept descriptions are build based on concept and role constructors, as well as names from $N_{C}$ , $N_{R}$ or $N_{I}$. If an individual name is used in a \emph{TBox} axiom, then it's called a nominal. \\
Different families of Description Logic provide different sets of constructors, besides axiom types. One of the simplest DLs is known as $\mathcal{AL}$ (Attributive Language). This DL supports concept conjunction ($C \sqcap D$, \texttt{owl:intersectionOf}), universal quantification ($\forall R.C$, \texttt{owl:allValuesFrom}), limited existential quantification ($\exists R.\top$, \texttt{owl:someValuesFrom} with a filler restricted to \texttt{owl:Thing}) and atomic negation ($\neg A$, \texttt{owl:complementOf}). More expressive DLs can be obtained from $\mathcal{AL}$ by adding further constructors. Each constructor is given a specific letter which is used to derive a name for any particular DL. For example, adding full negation ($ \neg C$) to $\mathcal{AL}$ produces the DL $\mathcal{ALC}$, which also contains concept disjunction ($C \sqcup D$, \texttt{owl:unionOf}) and full existential quantification ($\exists R.C$). However, extending the logic $\mathcal{ALC}$ with transitive roles becomes the logic $\mathcal{S}$. Then, $\mathcal{SH}$ extends $\mathcal{S}$ with role hierarchies $\mathcal{H}$ (\texttt{rdfs:subPropertyOf}). Adding nominals \textbf{$\mathcal{O}$} (\texttt{owl:oneOf}), inverse properties \textbf{$\mathcal{I}$} (\texttt{owl:inverseOf}) and number restrictions \textbf{$\mathcal{N}$} (\texttt{owl:minCardinality}, \texttt{owl:maxCardinality} or \texttt{owl:exactCardinality}) to $\mathcal{SH}$ produces $\mathcal{SHOIN}$. The latter one is the DL that underpins OWL 1. OWL 2 extends the expressivity of OWL 1 with qualified cardinality \textbf{$\mathcal{Q}$} to give $\mathcal{SHOIQ}$, and reflexive (\texttt{owl:ReflexiveProperty}), irreflexive (\texttt{owl:IrreflexiveProperty}), complex chains (\texttt{owl:propertyChainAxiom}) and disjoint properties \textbf{$\mathcal{R}$} to give $\mathcal{SROIQ}$. \\
Generally speaking, a concept in DL is referred to as a class in OWL. A role in DL is a property in OWL, which could be an \texttt{Object Property} (properties for which the value is an individual) or a \texttt{Data Property} (properties for which the value is a data literal). Axioms and individuals have the same meaning in DL and OWL. Owe to snugness   connection between OWL and DLs, in this paper, we will make no distinction between ontologies (in OWL) and knowledge bases (in DL)\footnote{In the remainer of this paper, by OWL we will mean OWL2.}.
\subsection{Reasoning}
At the crossroads of the ontology and Description Logic (DL) respective communities, there is a suite of inference services held to be the key of most applications or knowledge engineering efforts \cite{DLBook03}. These services are usually provided by automated decision systems. In DLs these systems, so-called \textit{reasoners}, implement decision procedures (for instance, the\emph{Tableau algorithm }\cite{BaaderTableau}, \emph{Hyper-Tableau} \cite{Motik09} and \emph{Conse\-{quence-Based}} \cite{Frantisek11}). They infer logical consequences from a set of explicitly asserted facts or axioms and typically provides automated support for reasoning tasks \cite{DLBook03}. These tasks are namely: \emph{satisfiability}, \emph{subsumption}, \emph{classification}, \emph{consistency} and \emph{realisation}.\\
Among these tasks, \emph{classification} is considered as the key reasoning task. It computes the full concept and role hierarchies. Explicit and implicit subsumption will be derived to help users navigating through the ontology towards mainly explanation and/or query answering respective tasks. Thus, it's supported by all modern DL reasoners and its duration is often used as a performance indicator to benchmark reasoning engines \cite{Abburu2012}. From an application point of view, an ontology should be classified regularly during its development and maintenance in order to detect undesired subsumptions as soon as possible. To make this feasible, in particular for large ontologies, classification has to be carried out as swiftly as possible. However, the increasing complexity of modern ontologies is an actual hamper towards reaching such a goal. Since OWL 2 is a highly expressive language, key reasoning tasks like consistency checking have an extremely high worst case complexity: 2NExpTime-Complete \cite{Horrocks06}, i.e., \emph{intractable}. However, this complexity could be tractable  with less expressive fragments of OWL 2, mostly known as OWL profiles like \emph{OWL EL}, \emph{OWL RL} and \emph{OWL QL} \footnote{For further reading about OWL 2 profiles, the reader is kindly referred to http://www.w3.org/TR/owl2-profiles/.}. Each profile limits the class, property and axiom constructors that OWL 2 admits, and consequently decreases the hardness level of the language. This restriction was decided in order to make it possible, and easy to implement efficient and scalable reasoners. However, the full power of the OWL 2 is still available under the OWL-DL profile. Reasoner designers are keeping optimizing their reasoning algorithms to overcome the complexity and the intractability of the latter profile.
\section{Related Works} \label{works}
In this section, we tried to draw out the landscape of the state of art works, which discussed the ontology complexity  key features, likely to impact reasoner performances. These works were graded into three main categories considering their investigation scopes: first, works assessing the ontology quality which introduced to the community a huge amount of ontology metrics; then works evaluating reasoners quality in terms of computational time; and finally works attempting to correlate reasoner empirical behaviours to particular ontology features. Table \ref{tab:relatedworks} sum ups the main aspects of the aforementioned categories. In the following, we give a more detailed review of works falling in these categories.\\

\begin{table*}
\centering
\caption{The landscape of tools and methods about ontology features altering reasoner performances}
\setlength{\tabcolsep}{0.1cm}
\begin{tabular}{|p{1.5cm}|p{2cm}|p{2cm}|p{2cm}|p{4cm}|} 
\hline 
\textbf{Scope} & \textbf{Purpose} & \textbf{References}&\textbf{Ontology Design}&\textbf{Ontology features}  \\
\hline
Ontology & Quality evaluation & \cite{LePrendu10,theevaluation07,ontoQA,Zhang10} & Graph 
/ OWL / RDF & Structural, Syntactic, Semantic, etc \\ 
\hline 
Reasoner & Performance evaluation & \cite{Abburu2012,ORe2013Results,ORE2014} & KB & Size + Expressivity \\ 
\hline
 & Ontology  &\emph{Tweezers} \cite{WangP07} &OWL & Patterns \\ 
 \cline{3-5} 
Reasoners &Profiling & \emph{Pellint} \cite{Pellint08}& OWL & Patterns \\ 
 \cline{2-5} 
Empirical& Reasoner& {\small Gon\c{c}alves et al. }\cite{Hotspots12}& KB & SAT runtime \\
 \cline{3-5} 
behaviours& improvement & {\small Romero et al.} \cite{More12} & KB & Expressivity 
\\  \cline{2-5} 
 \texttt{w.r.t. } &Reasoner & Kang et al. \cite{Predicting12} &Graph  & 27 features filtered via feature selection algorithms \\ 
 \cline{3-5} 
Ontology &Performances & Sazonau et al. \cite{PredictingLocGlob14}&OWL & 57 features filtered with PCA technique\\ 
 \cline{3-5} 
Features &Prediction & Kang et al. \cite{Predict2014} &Graph + OWL & 91 features filtered based on correlation removal. \\ 
\hline
\end{tabular} \label{tab:relatedworks}
\end{table*}

Ontology knowledge richness and conceptualization quality is widely assessed in the literature. Huge stream of ontology evaluation metrics was proposed for this purpose \cite{theevaluation07,ontoQA,LePrendu10}. However, little attention was paid to investigate the effectiveness of these metrics to assess the hardness of ontologies against reasoning tasks \cite{Zhang10}. In the other hand, reasoner benchmarks and competitions \cite{Abburu2012,ORe2013Results,ORE2014} are annually held to compare the latest innovations in the field of semantic reasoning. The performances of these engines against well selected ontologies are evaluated mainly considering the computational runtime. Roughly speaking, the reasoner performances depend on the success or the failure of optimizations tricks set up by reasoner designers to overcome particular known DL complexity sources. However, theses tricks would lead to enormous performance variability across the inputs which is still hardly predictable a priori. Yet, it is not well understood which particular aspect in the ontology is lowering the robustness of reasoners, besides the usual reported features mainly the ontology size and the expressivity. Recently, some tools, e.g.  \emph{Tweezers} \cite{WangP07} and \emph{Pellint} \cite{Pellint08}, tried to give insights about reasoner performances bottlenecks \texttt{w.r.t.} the input ontology. To fulfil this task, software profiling techniques were deployed. The first tool reports particular performance statistics\footnote{These statistics are general ones like the runtime and the memory occupation and, more detailed ones focusing on the behavior of the reasoning algorithm like the size of completion graph and the number of steps to find a clash \cite{DLBook03}.} of the satisfiability task (SAT) processed by the reasoner \emph{Pellet} \cite{Pellet07}. In addition, authors of the tool have reported that some of this reasoner performance bottlenecks are caused by particular modelling patterns used in the input ontology. Similarly, \emph{Pellint} examines the ontology to report and even to repair some ontology modelling pitfalls. These pattern like structures were suspected to brook down the \emph{Pellet}'s runtime. Despite, the worth of these proposals, it's hard to agree on their effectiveness, since they were proposed considering one particular reasoner. However, reasoning methodologies vary from one engine to another and there is no agreement that they share the same bottlenecks. More recently, Gon\c{c}alves et al. \cite{Hotspots12} suggested that there are ontologies which are \emph{performance homogeneous} and \emph{performance heterogeneous} ones. The heterogeneity is clause to particular entanglements between axioms in the ontology, causing the increase of the reasoning cost. Authors proposed a method to track these entanglements and extract their corresponding ontology modules. The latter ones were called ontology \emph{Hotspots}. They also introduced a method to approximate reasoning with the Hotspots. However, their experiments have revelled that there is no precise co-relation between the reasoning time of a \emph{hotspot}
alone, and the reduction in reasoning time when such \emph{hotspot} is removed. They affirmed that more investigations should be made about possible interactions between the hotspots and other ontology features. A further elaborated method to boost the reasoning using modularization techniques was introduced by Romero et al. \cite{More12}. The proposed reasoner \emph{MoRe} uses the ontology expressivity as a partitioning criteria aiming at extracting a relatively "easy-to-handle" module and a hard one. Two reasoners are then coupled, each of which known for its appropriateness for the extracted modules.\\

Another steam of works, mainly described in \cite{Predicting12,PredictingLocGlob14,Predict2014}, used supervised machine learning techniques aiming at predicting the computational time of a reasoner for a given ontology. Their predictive models take advantage from a large set of pre-computed ontological metrics.  The rational behind this choice is to be able to automatically learn future reasoner's behaviours based on what was experienced in their previous executions. Kang et al. \cite{Predicting12} were the first to apply machine learning techniques to predict the ontology classification computational time carried by a specific reasoner. 27 metrics were computed for each ontology. These metrics were previously proposed by a work stressing on ontology design complexity \cite{Zhang10}. The labels to be predicted were time bins specified by the authors. They learned random forest-based models for 6 state of art reasoners and obtained high accuracy values. Moreover, they proposed an algorithm to compute the impact factors of ontology metrics according to their effectiveness in predicting classification performances for the different reasoners. Kang et al. have further improved their approach, in a more recent work \cite{Predict2014}. They replaced time bins labels by concrete values of reasoner classification runtime and proposed additional metrics. They also demonstrated the strengths of their predictive models by applying them to the problem of identifying ontology performance \emph{Hotspots}. On the other hand, Sazonau et al. \cite{PredictingLocGlob14} claimed that Kang's metrics based on graph translation of OWL ontologies are not effective. Thus, they proposed another set of metrics and used more machine learning techniques to reduce the dimensionality of the ontology feature vector in order to identify the key features, likely to correlate the most with the reasoning performances.\\

Clearly, machine learning methods proposed in the last steam of works are the closest to meet our needs. Indeed, the impact of particular ontology features on reasoner performances are automatically investigated using empirical knowledge about reasoners. These methods are generic enough, that they would be applied to any reasoner, with the only requirement to provide enough running results of this reasoner on diverse ontologies. Nevertheless, choosing good features is crucial to the construction of good predictive models. Unfortunately, our review of state of art confirmed that there is no known, automatic way of constructing good ontology feature sets. Instead, we believe that we must use distinct domain knowledge to identify properties of ontologies that appear likely to provide useful information. Afterwards, applying supervised machine learning techniques would be appropriate to examine the real impact of these features on reasoning performances and help selecting the most relevant ones.
\section{Ontology features altering the reasoner performances} 
\label{ontofeatures}
A wealthy number of ontological  features was introduced in literature, particularly to build learning models for reasoner computational time prediction. We reused some of them and defined new ones, that we thought could be relevant to evaluate the empirical hardness of reasoners. Mainly, we discarded those computed based on specific graph translation of the OWL ontology. In fact, Sazonau et al. \cite{Predict2014} have previously argued that these kind of features are not reliable as there is no agreement of the way an ontology should be translated into a graph. 

We split the ontology features into 4 categories: (\textit{i}) size description; (\textit{ii}) expressivity description; (\textit{iii}) structural features; and (\textit{iv}) syntactic features. Within these categories, features are intended to characterize specific aspect of the ontology design. The third and fourth category are further split into subcategories that provide a finer description of the ontology content. In overall, 112 ontology feature was characterized, which will be described in the next sections:
\subsection{Ontology Size Description}  \label{ontosize}
To characterize the size of the ontology, we propose 6 features, explained in the following: 
\begin{description}[leftmargin=*,parsep=0cm,itemsep=0cm,topsep=0cm]
\item[Signature size features]: we design 5 features to assess the amount of terms defined in an ontology. Given an ontology signature $\widetilde{\mathcal{O}}=\langle N_{C}, N_{R}, N_{I} \rangle$, we count the number of names $nc_i$, that uniquely identify classes in the ontology, $nc_i \in N_{C}$.  We call this feature \textbf{SC}, the size of the ontology  classes, where SC=$|N_{C}|$. Analogously, we compute the number of user-defined object and data property names, respectively denoted by \textbf{SOP} and \textbf{SDP}, where $|N_{R}|=SOP+SDP$, then the number of named individuals \textbf{SI}=$|N_{I}|$. In addition, we record the number of data types\footnote{These are RDF literals or simple types defined in accordance with XML Schema datatypes.} defined in the ontology \textbf{SDT}.
\item[Axioms size features (OAS)]: As commonly known, reasoners only deal with axioms belonging to the subsets \emph{TBox}, \emph{RBox} or \emph{ABox}. Annotations are simply ignored when processing an ontology for a reasoning task. Therefore, we distinguished between two features, \textbf{(SLA)} which stands for the number of OWL axioms qualified as logical ones and (\textbf{SA}) which designs the total number of axioms in the ontology.
\end{description}
Worth of mention, our proposed set of signature size features is a particular case of \textbf{SOV}, size of vocabulary, introduced by Zhang et al.\cite{Zhang10}. While, the latter one opted for measuring to measure the complexity of an ontology by simply counting the total number of its named entities, we rather preferred to distinguish between these entities and compute separately each of them. 
\subsection{Ontology Expressivity Description}  \label{ontoexpressivity}
In Section \ref{background}, we recalled basic elements of the OWL vocabulary, the DL families they belong to and then, we pointed out how the worst case complexity of reasoning tasks are closely depending on the expressive level of the ontology language. We retained two main features to identify the expressivity of the ontology language, namely:
\begin{description}[leftmargin=*,parsep=0cm,itemsep=0cm,topsep=0cm]
\item [OWL profile name (\textbf{OPR})]: as above mentioned, there are four possible profiles DL, EL, QL and RL. We record the one the ontology language fits in. However, in some particular cases, the vocabulary and language constructs used in an ontology may violate rules of all the OWL profiles. In this case, the profile of the ontology is denoted by a virtual profile name, that we called PNAN. In contrast, we tag by PFULL  the ontology  that matches all the OWL profiles. 
\item [DL family name (\textbf{DFN})]: it is a more strict denomination of the DL constructs group used in the ontology. For instance, an ontology could be $\mathcal{AL}$ or  $\mathcal{ALC}$ or  $\mathcal{SHOIN}$, etc.  Basically, each ontology has a unique DL family name.
\end{description}
\subsection{Ontology Structural Description}  \label{ontostruture} 
We paid a special attention to characterize the taxonomic structure of an ontology, i.e., its inheritance hierarchy. The latter sketches the tree like structure of  subsumption relationships between names classes $A \sqsubseteq  B$ or named properties $R \sqsubseteq S$. We remind that a reasoner classification task infer implicit subsumption from the explicitly defined ones. So, the more the inheritance hierarchy is complex and over-sized, the more the reasoning computational time may be important. In this category, we gathered various features that have been defined in literature to describe concept hierarchies. These are, basically, metrics widely used by ontology quality evaluation community \cite{Gangemi06,ontoQA,LePrendu10}. The following subcategories describe the essence of the retained features. 
\subsubsection{Class and property hierarchical features}
We build both concept hierarchy denoted by \textbf{Chierarchy} and property hierarchy denoted by \textbf{PHierarchy}. Interestingly enough, only subsumption relations between named object properties specified by the axiom \texttt{owl:subPropertyOf} were considered for the \textbf{HP} construction. It means that property characteristics like inverse, transitivity, reflexivity and symmetry were ignored. Then for each hierarchy, we measured the following features:
\begin{itemize}[leftmargin=*,parsep=0cm,itemsep=0cm,topsep=0cm]
\item \textbf{C\_MD}, \textbf{P\_MD} it is the maximal depth of class and property respective hierarchies. This feature was identified by LePendu et al. \cite{LePrendu10} as a possible reasoning complexity source.
\item \textbf{C(P)\_MSB}, \textbf{C(P)\_ASB}: the maximal and the average number of subclasses (resp. sub-properties) in a class (property) hierarchy. This feature was called by Tartir et al. as \emph{Inheritance Richness} \cite{ontoQA}. The authors claimed that higher values of this feature would indicate that the ontology is deep. However, lower value would lead to a shallow horizontal ontology having less detailed knowledge.
\item \textbf{C(P)\_Tangledness}, \textbf{C(P)\_MTangledness}: \emph{tangledness} is owe to Gangemi et al. \cite{Gangemi06} and measures the number of classes in an ontology with multiple superclasses. We computed this feature for both class and property hierarchy and we also recorded the maximal number of named superclasses (\textbf{C(P)\_MTangledness}). Tangledness was also called \emph{tree impurity} by Zhang et al. \cite{Zhang10}. Then, Kang et al. \cite{Predicting12} have highlighted, that this feature has a worth of cite impact factor on reasoner performances.
\end{itemize}
\subsubsection{Cohesion features} 
The literature provides a plethora of various metrics to design the \emph{Cohesion} of the ontology, otherwise the degree of relatedness of its entities. We retained the ones introduced by Faezeh and Weichang \cite{Cohesion10}.
\begin{itemize}[leftmargin=*,parsep=0cm,itemsep=0cm,topsep=0cm]
\item \textbf{CCOH}, \textbf{PCOH}: these are respectively class hierarchy cohesion and property hierarchy cohesion. They are based on the number of direct and indirect hierarchical links. We report the used formula to compute the class hierarchy cohesion:
\begin{equation}
CCOH= \frac{2\times(NdHC+NidHC)}{NC^{2}-NC}
\end{equation}
where, $NdHC$ is the number of direct hierarchical links between classes, $NidHC$ is the indirect ones and $NC$ is number of named classes in hierarchy. \textbf{PCOH} is computed in the same way where classes are replaced by properties.
\item \textbf{OPCOH}: it is the object property cohesion. This feature is computed using the number of classes which have been associated through the particular object property (domain and range). Given an object property $op_{i}$, the number of classes in its domain $NdC(op_{i})$ and the number of classes in its range $NrC(op_{i})$, i.e., 
\begin{equation}
OPCOH= \frac{ 2 \times \sum_{i=1}^{NOProp} NdC(op_{i}) \times NrC(op_{i}) }{ NOProp \times (NC^{2}-NC)}
\end{equation}
\item \textbf{OCOH}: the ontology cohesion is simply a weighted aggregation of the previously defined cohesion metrics, i.e, \textbf{CCOH}, \textbf{PCOH} and the \textbf{OPCOH}.
\end{itemize}
\subsubsection{Schema Richness features} 
Finally, we enriched the ontology structural category by two additional features proposed in Tartir et al. \cite{ontoQA}. These features are well known for ontology evaluation community as they are part of the \emph{OntoQA} tool.
\begin{itemize}[leftmargin=*,parsep=0cm,itemsep=0cm,topsep=0cm]
\item \textbf{RRichness}: Relationship richness reflects the diversity of relations in ontology. Formally, it is the ratio of the number of relations between classes that are not hierarchical \texttt{w.r.t.} total number of different types of the ontology relationships. We slightly modified the formal expression of this metric in order to be able to compute it without any translation of ontology into a graph.
\item \textbf{AttrRichness}: The attribute richness is  defined as the average number of attributes per class. To compute this feature, we considered data properties as class attributes.
\end{itemize}
We discarded other metrics defined by Tartir et al., as they overlap with those that we have already suggested in other categories.
\subsection{Ontology Syntactic Features} \label{semdimensio}
Our main purpose when collecting features for this category, is to quantify some of the general theoretical knowledge about DL complexity sources that would degrade reasoner performances and eventually lead to unexpected reasoning results. To accomplish this purpose, we conducted an investigation about main reasoning algorithms \cite{BaaderTableau,Motik09}. Thus, we gathered relevant ontology features, that have inspired the implementation of well known reasoning optimization techniques \cite{ChapterIan2003,TsHP07}. Features of the current group are divided in 6 subcategories, covering different aspects of the ontology syntactic elements. This organization was inspired by the definition of feature levels provided by Kang et al. \cite{Predicting12}.
\subsubsection{Features of Axioms level}  
Reasoner process differently each type of axiom with different computational cost \cite{DLBook03}. In this category, we gathered features that attempt to characterize the different types of axioms as well as to assess their respective relevance in the ontology.
\begin{itemize}[leftmargin=*,parsep=0cm,itemsep=0cm,topsep=0cm]
\item \textbf{KB sub-parts features (KBF)}:  in Section \ref{ontology}, we recalled that a knowledge base (KB), which is in our case the ontology, has three main parts \emph{TBox}, \emph{RBox} and \emph{ABox}, each of which has a specific set of axioms, otherwise each have a specific size. Given this description, we recorded the size ratio of these KB subsets \texttt{w.r.t.} the ontology axiom size (\textbf{OAS}) and we denoted them \textbf{RTBx}, \textbf{RRBx}, \textbf{RABx}.
\item \textbf{Axiom Types Frequencies (ATF)}: this is a set of 28 features, each of which corresponds to a particular OWL axiom type. In Section \ref{ontology}, we remind some of the axiom types described in the OWL official specification, like \texttt{rdfs:\-{subClassOf}}, \texttt{rdfs:subPropertyOf}, etc. By frequency we mean, the ratio between the number of occurrences of a given axiom type, and the ontology axioms size (\emph{OAS}).
\item \textbf{Axiom Depth Feature}: we compute the maximal parsing depth of axioms in the ontology \textbf{AMP}. It is a feature of common use, described by Sazonau et al \cite{Predict2014}. An axiom depth is the number of this axiom levels of nested expressions. This feature attempts to capture the extend of structural complexity of an axiom in a given ontology. We added to this information, the average nesting depth \textbf{AAP} of all axioms, in order to give insight about how common is this type of complexity in the ontology.
\end{itemize}  
\subsubsection{Features of Constructors level} 
In Section \ref{background}, we remind that DLs can be classified in different classes of expressiveness, depending on the constructors they provide. This expressiveness has a general impact on the computational complexity of reasoning tasks performed on the ontology.  In previous reasoner prediction works \cite{Predicting12,Predict2014}, authors simply counted axioms that involve potentially hard constructors. However, they missed that one constructor could be invokeds more than once in the same axiom. Moreover, we believe that the \emph{density} of use of constructors, may be a valuable indicator of the ontology complexity. In order to characterize these informations, we propose three features that we describe in the following:
\begin{itemize}[leftmargin=*,parsep=0cm,itemsep=0cm,topsep=0cm]
\item \textbf{Class Constructors Frequencies} (\textbf{CCF}):  this is a set of 11 features, where each element is a specific constructor frequency in the ontology. Formally, given a class constructor $cc_{i}$ belonging to the set of all OWL class constructors ($cc_{i} \in CC$), \textbf{CCR($cc_{i}$)} is defined as the ratio of the $cc_{i}$ total occurrences in each \emph{TBox} axiom ($A_{tx} \in \mathcal{T}$), divided by the sum of all constructors occurrences. The value of a \textbf{CCR} feature ranges within the unit interval $[0,1]$, i.e,
\begin{equation}
CCR(cc_{i})=\frac{\sum_{j=1}^{\vert \mathcal{T} \vert} Count(cc_{i}, A_{tx_{j}})}{ \sum_{i=1}^{\vert CC \vert} \sum_{j=1}^{\vert \mathcal{T} \vert} Count(cc_{i}, A_{tx_{j}})}, cc_{i} \in CC.
\end{equation}
\item \textbf{Ontology Class Constructors Density (OCCD)}: we proposed another feature to compute the overall constructors \emph{"density"} in the ontology. Formally, it computes the ratio of the total number of all constructors occurrences, divided by the maximal possible number of constructors in the ontology. The latter is defined as the multiplication result of the total number of \emph{TBox} axioms $|\mathcal{T}|$, by the maximal counted number of constructors in one \emph{TBox} axiom. The value of \textbf{OCCD} ranges within the unit interval $[0,1]$, i.e,
\begin{equation}
OCCD =\frac{\sum_{i=1}^{\vert \mathcal{T} \vert} \sum_{j=1}^{\vert CC \vert} Count(cc_{j}, A_{tx_{i}})}{ |\mathcal{T}| \times max (\sum_{j=1}^{\vert CC \vert} Count(cc_{j}, A_{tx}), \forall A_{tx} \in \mathcal{T})}.
\end{equation}
\item \textbf{Constructors Coupling Patterns (CCP)}: we also defined a more sophisticated feature that examines particular combinations of constructors, that may increase the inference computational cost, whenever used in an axiom. The rational behind this proposition comes from lectures about the well known \emph{Tableau} algorithm \cite{BaaderTableau}. In fact, while checking a satisfiability of a concept by the latter algorithm, expansion rules are recursively applied in order to build a completion graph, called the \emph{model}. Each class constructor has its own expansion rule. Applied in a specific order, the rules may lead to a sharp increase of the completion graph size, and hence the reasoning cost. We have specified three particular patterns, which describe fragments with \textit{"costly"} class constructors combinations. We detected the occurrences of each of these patterns by \emph{SPARQL} \footnote{The specification of SPARQL query language is available at http://www.w3.org/TR/sparql11-query/} based queries, that we have written for this purpose. The \emph{CCP} set of patterns are described in the following. For each pattern, we recorded its occurrences in the ontology.
\begin{itemize}[leftmargin=*,parsep=0cm,itemsep=0cm,topsep=0cm]
\item  \textbf{IU} \emph{(Intersection, union Pattern)}: an IU pattern is reported  when a conjunction (\texttt{owl:intersectionOf}) of class expressions in an axiom is part of a disjunction (\texttt{owl:unionOf}) of class expressions in the same axiom, and vice versa. This pattern can be manifested by one of the following forms:
\begin{align}
\sqcap\quad(\ldots,\sqcup(C_{1}, C_{2}, \ldots), \ldots) \\
\sqcup\quad(\ldots,\sqcap(C_{1}, C_{2}, \ldots), \ldots ) 
\end{align}
\item \textbf{EUvI} \emph{(Existential, universal having intersection Pattern)}: an EUI pattern occurrence is defined by a conjunction of class expressions, that concurrently contains an existential restriction and a universal restriction associated to the same role "$r$". This pattern can be manifested by one of the following forms:
\begin{align}
\sqcap\quad(\ldots,\exists r.C, \forall r.D, \ldots) \quad  \\
C_{1} \sqsubseteq \exists r.C \quad and \quad C_{1} \sqsubseteq \forall r.D
\end{align}
\item \textbf{CUvI} \emph{(Cardinality, universal having intersection Pattern)}: The CUvI pattern is a particular case of the EUvI pattern, where existential restriction is replaced by a some restriction forms ($\leq n r.C$, $\geq n r.C$, $= n r.C$).
\end{itemize}
\end{itemize}
\subsubsection{Features of Class level}
Classes in the ontology could be named or specified via complex expressions. In this subcategory, we will highlight different methods to define classes and track their impact in the ontology \emph{TBox} part. 
\begin{itemize}[leftmargin=*,parsep=0cm,itemsep=0cm,topsep=0cm]
\item \textbf{Class Definition Features (CDF)}: in \cite{ChapterIan2003},  Horrocks established that restricting the Knowledge base (KB), $\mathcal{K}= \langle \mathcal{T}, \mathcal{R}, \mathcal{A} \rangle$  to unique and acyclic definition axioms, makes reasoning much easier, as the \emph{unfolding} technique could be applied to all axioms. Concept definition axioms are primitive ones \textbf{PCD} of the form $A\sqsubseteq D$, or non primitive ones \textbf{NPCD} of the form $A \equiv D$ , where $A$ is an atomic concept name. However, real-world KBs commonly contain general definition axioms \textbf{GCI}. These  axioms are of the form $C \sqsubseteq D$  or $C \equiv D$ , where both $C$ and $D$ are complex class descriptions. They are known to be costly to  reason with, due to the high degree of non-determinism that they introduce. Thus, optimization techniques, mainly \emph{Absorption}, is commonly used to reduce the number of \textbf{GCI} in the ontology by manipulating them to have the form of a primitive concept definition \textbf{PCD}. \emph{Absorption} is widely implemented in DL reasoners, and designers are often looking to increase its optimization power. Motivated enough, we proposed to record the ratio of each of these kind of class definitions (PCD, NPCD, GCI) \texttt{w.r.t.} \emph{TBox} size.
\item \textbf{Cyclic Class Feature (CCyc)}: we computed cyclic class definitions in the ontology and retained their ratio  \texttt{w.r.t.} total number of named classes, i.e, the \textbf{SC} feature. In DL, a cyclic definition axiom is the one that references the same (or equivalent) classes (or properties) on both sides of the subsumption relation (i.e $\exists P.C \sqsubseteq C$, or $P \circ P \sqsubseteq P$). Such an axiom may be explicit of inferred by a reasoner. We only computed the explicit ones by implementing the method described by Baader et al. \cite{Cycle2011}.
\item \textbf{Class Disjointness Feature (CDIJ)}: this feature stands for the ratio of named classes declared as disjointed \texttt{w.r.t.} the class size \textbf{SC}. Our motivation to compute this feature is based on stated observations by Wang et al. \cite{WangP07}. The latter has conducted empirical studies on reasoners using his profiling tool \emph{Tweezers}. They highlighted that there is a crucial need to characterize the "\emph{right amount"} of disjointness statements to be put in an ontology, as some of them can greatly reduce the computational time when inferring an ontology, but also too many statements would remarkably increase the runtime. The class disjointness is also examined by the \emph{Pellint} tool \cite{Pellint08}, and reported as a bottleneck when the number of statements exceeds some predefined threshold. 
\item \textbf{Class Form Nominals (CNOM)}: this feature record the ratio of classes defined based on named nominals \texttt{w.r.t.} total number of named classes. 
\end{itemize}
\subsubsection{Features of Properties level}
Interestingly enough, we will characterize special features of the ontology properties, in particular, the object property.
\begin{itemize}[leftmargin=*,parsep=0cm,itemsep=0cm,topsep=0cm]
\item \textbf{Object Property Characteristics Frequencies (OPCF)}: this is a set of 9 features in relation with particular object property characteristics. In OWL, the latter ones are defined using specific axioms that describe object properties transitivity, symmetry,  reflexivity and etc. Horrocks and Tsarkov \cite{ChapterIan2003,TsHP07} have respectively emphasized 
on the hardness of managing particular object property description characteristics, since they impact the effectiveness of some reasoning optimization techniques \footnote{For example, \emph{Internalisation} and \emph{Caching} are less effective at the presence of inverse properties in the ontology}. Hence, more sophisticated and probably costly reasoning procedures would be required to overcome these characteristics hardness level. We denoted the set of all object property characteristics as $OPC$ and we defined $OPCF$ as an object property characteristic frequency. To made clear, when specifying an object property as transitive, only one axiom is required (\texttt{owl:TransitiveObjectProperty}). However, this object property name could be repeatedly involved in many other \emph{TBox} axioms and even more than once in one axiom. Consequently, reasoning techniques dealing with transitivity would be applied as far as this transitive object property is used. Formally, to compute $OPCF$ of a given object property characteristic $C^{op}_{i} \in OPC$, we start by collecting named object properties having $C^{op}_{i}$ in common. We designed the latter set as $S(C^{op}_{i})=\{OP_{j}, j\geq 0\}$. Then, we sum up the occurrence value of each element in this collection and we denoted it $OPCO(C^{op}_{i})$. Later, this value is divided by the sum of the total characteristic occurrences, hence we obtain an $OPCF(C^{op}_{i})$ value ranging in $[0,1]$. By computing this latter ratio, we would be able to flag out which of the object properties characteristic have the highest impact on the reasoning process.
\begin{equation}
OPCO(C^{OP}_{i})= \sum_{j=1}^{\vert S(C^{op}_{i}) \vert} \sum_{k=1}^{\vert \mathcal{T} \vert} Count(OP_{j}, A_{tx_{k}})
\end{equation}
\begin{equation}
OPCF (C^{OP}_{i})= \frac{OPCO(C^{OP}_{i})}{\sum_{j=1}^{\vert OPC \vert} OPCO(C^{OP}_{j}) }
\end{equation}
\item \textbf{Number Restriction Features (HVC, AVR)}: we studied the impact of using high values with object properties number restrictions. So, we retained for each cardinality type that is min, max and exact cardinality, its highest respective values. For instance, taking a restriction of the form $\geq nR.C$, we have recorded $max(n)$ of all restrictions having the same form. Thus, we build the set of highest values of cardinalities, and we denoted it (\textbf{HVC}). In addition, we computed the average value of used numbers for cardinality restrictions, and we denoted it (\textbf{AVC}). Worth of cite, the \emph{Pellint} tool \cite{Pellint08} reports an ontology pitfall when cardinality values exceed some predefined threshold. However, it's not known how this threshold is fixed.
\end{itemize}
\begin{figure*}
\centering
\includegraphics[scale=0.75]{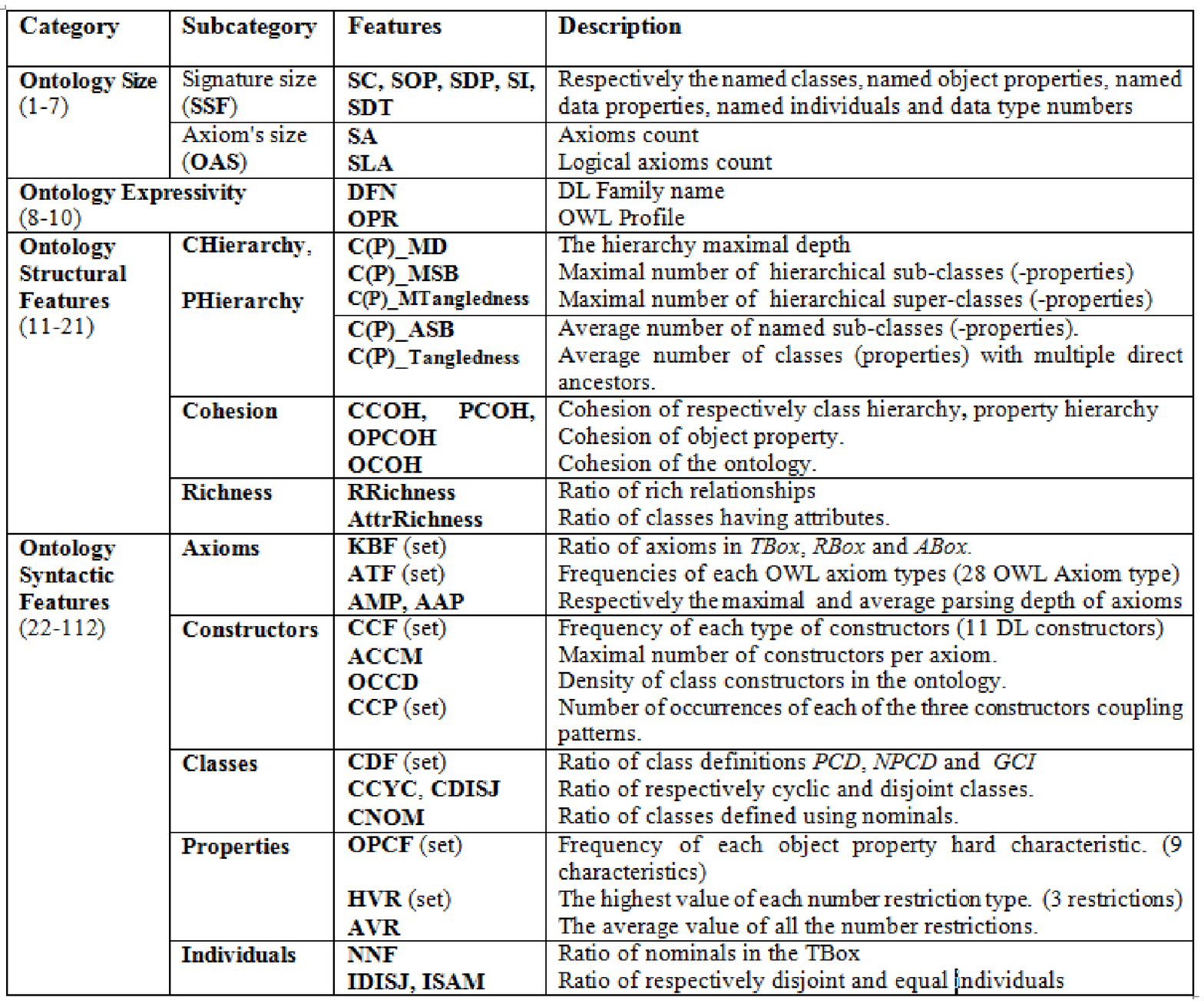}
\caption{The ontology features catalogue}
\label{fig:featuresdesc}
\end{figure*}
\subsubsection{Individual level Features} In this subcategory, we specify some of the interesting characteristics of named individuals that would be declared in the ontology.
\begin{itemize}[leftmargin=*,parsep=0cm,itemsep=0cm,topsep=0cm]
\item \textbf{Nominal Frequency Features (NomTB, TBNom)}: in the remainder Section \ref{background}, we mentioned that named individuals could be used in \emph{TBox} axioms to define new classes. In this case, individuals are designed as nominals. However, this modeling method come at a price, since nominals require specific reasoning procedures that would lower the runtime of the reasoner \cite{TsHP07}. To capture the impact of using nominals, we counted their occurrences in \emph{TBox} axioms and retained its ratio (\textbf{NomTB}) \texttt{w.r.t.} the individuals size \textbf{SI}. In addition, we recorded \textbf{TBNom}, the ratio of axioms having nominals \texttt{w.r.t.} \emph{TBox} size (\textbf{STBx}).
\item \textbf{Individual Similarity Features (IDISJ, ISAM)}: it is simply the ratio of named individuals defined as disjoint ones \textbf{IDISJ}, as well as the \textbf{ISAM} ratio of individuals declared as equal ones (\texttt{owl:sameAS}).
\end{itemize}
\section{Conclusion}\label{conclusion}
In this report, we investigated key ontology features, likely to impact reasoner performances. First, we reviewed state of art works, which highlighted the variability of reasoner empirical performances. We outlined the previous efforts to correlate between particular ontology features and reasoner performances. Then, we introduced a large set of comprehensive features covering different aspects of the ontology structural and syntactic characteristics. These features were split up into 4 main categories, some of them are further brook down into subcategories, to capture more finer description of the ontology components. Throughout our study, we tried to quantify theoretical and empirical knowledge about the ontology complexity sources. We believe that we gathered concise and rich set of ontology features, likely to be good indicators of its hardness level against reasoning tasks. \\
Our features could be used in any machine learning process. Thus for future works, we planed to conduct a supervised learning study based aiming to predict reasoner empirical behaviours based on our ontology features. We believe that this technique would help us unveil the key ontology features, with respect to the reasoning task. Gaining insights about which makes the ontologies hard to process, would help improving both their modelling, and revision process, throughout avoiding or repairing the hard features.
\bibliographystyle{splncs03}
\bibliography{bib_recherche}

\begin{thebibliography}{10}
\providecommand{\url}[1]{\texttt{#1}}
\providecommand{\urlprefix}{URL }

\bibitem{Abburu2012}
Abburu, S.: A survey on ontology reasoners and comparison. International
  Journal of Computer Applications  57(17),  33--39 (2012)

\bibitem{Cycle2011}
{Baader}, F., {Borgwardt}, S., {Morawska}, B.: Unification in the description
  logic {$\mathcal{EL}$} w.r.t.\ cycle-restricted {TB}oxes. Ltcs-report,
  Institute for Theoretical Computer Science, Technische Universit{\"a}t
  Dresden, Germany (2011)

\bibitem{DLBook03}
Baader, F., Calvanese, D., McGuinness, D.L., Nardi, D., Patel-Schneider, P.F.
  (eds.): The Description Logic Handbook: Theory, Implementation, and
  Applications. Cambridge University Press, USA (2003)

\bibitem{BaaderTableau}
Baader, F., Sattler, U.: Tableau algorithms for description logics. In:
  Proceedings of the International Conference on Automated Reasoning with
  Analytic Tableaux and Related Methods (TABLEAUX) (2000)

\bibitem{ORE2014}
Bail, S., Glimm, B., Jiménez-Ruiz, E., Matentzoglu, N., Parsia, B.,
  Steigmiller, A.: Summary ore 2014 competition. In: the 3rd Int. Workshop on
  OWL Reasoner Evaluation (ORE 2014), Vienna, Austria (2014)

\bibitem{Cohesion10}
Faezeh, E., Weichang, D.: Canadian semantic web. chap. A Modular Approach to
  Scalable Ontology Development, pp. 79--103. Springer US (2010)

\bibitem{Gangemi06}
Gangemi, A., Catenacci, C., Ciaramita, M., Lehmann, J.: Modelling ontology
  evaluation and validation. In: Proceedings of the 3rd European Semantic Web
  Conference (2006)

\bibitem{Hermit12}
Glimm, B., Horrocks, I., Motik, B., Shearer, R., Stoilos, G.: A novel approach
  to ontology classification. Web Semant.  14,  84--101 (2012)

\bibitem{ORe2013Results}
Gon\c{c}alves, R.S., Bail, S., Jim{\'e}nez-Ruiz, E., Matentzoglu, N., Parsia,
  B., Glimm, B., Kazakov, Y.: Owl reasoner evaluation (ore) workshop 2013
  results: Short report. In: ORE. pp. 1--18 (2013)

\bibitem{Hotspots12}
Gon\c{c}alves, R.S., Parsia, B., Sattler, U.: Performance heterogeneity and
  approximate reasoning in description logic ontologies. In: Proceedings of the
  11th International Conference on The Semantic Web. pp. 82--98 (2012)

\bibitem{ChapterIan2003}
Horrocks, I.: Implementation and optimisation techniques. In: The Description
  Logic Handbook: Theory, Implementation, and Applications, chap.~9, pp.
  306--346. Cambridge University Press (2003)

\bibitem{Horrocks06}
Horrocks, I., Kutz, O., Sattler, U.: The even more irresistible
  $\mathcal{SROIQ}$. In: Proceedings of the 23rd Benelux Conference on
  Artificial Intelligence. pp. 57--67 (2006)

\bibitem{Predicting12}
Kang, Y.B., Li, Y.F., Krishnaswamy, S.: Predicting reasoning performance using
  ontology metrics. In: Proceedings of the 11th International Conference on The
  Semantic Web. pp. 198--214 (2012)

\bibitem{Predict2014}
Kang, Y.B., Li, Y.F., Krishnaswamy, S.: How long will it take? accurate
  prediction of ontology reasoning performance. In: Proceedings of
  Twenty-Eighth AAAI Conference on Artificial Intelligence. pp. 80--86 (2014)

\bibitem{LePrendu10}
LePendu, P., Noy, N., Jonquet, C., Alexander, P., Shah, N., Musen, M.: Optimize
  first, buy later: Analyzing metrics to ramp-up very large knowledge bases.
  In: Proceedings of The International Semantic Web Conference. pp. 486--501.
  Springer (2010)

\bibitem{Pellint08}
Lin, H., Sirin, E.: Pellint - a performance lint tool for pellet. In:
  Proceedings of the OWL Experiences and Directions Workshop at ISWC'08. vol.
  432. Germany (2008)

\bibitem{Towards11}
Martin-Recuerda, F., Walther, D.: Towards understanding reasoning complexity in
  practice. In: Proceedings of Third International Conference of Knowledge
  Representation and Reasoning. UK (2011)

\bibitem{Motik09}
Motik, B., Shearer, R., Horrocks, I.: Hypertableau reasoning for description
  logics. Journal of Artificial Intelligence Research  36,  165--228 (2009)

\bibitem{theevaluation07}
Obrst, L., Ashpole, B., Ceusters, W., Mani, I., Smith, B.: Semantic web. chap.
  The evaluation of ontologies - Toward Improved Semantic Interoperability, pp.
  139--158. Springer US (2007)

\bibitem{More12}
Romero, A.A., Grau, B.C., Horrocks, I.: More: Modular combination of owl
  reasoners for ontology classification. In: Proceedings of the 11th
  International Conference on The Semantic Web (2012)

\bibitem{PredictingLocGlob14}
Sazonau, V., Sattler, U., Brown, G.: Predicting performance of owl reasoners:
  Locally or globally? In: Proceedings of the Fourteenth International
  Conference on Principles of Knowledge Representation and Reasoning (2014)

\bibitem{Frantisek11}
Simancik, F., Kazakov, Y., Horrocks, I.: Consequence based reasoning beyond
  horn ontologies. In: Proceedings of the 22nd International Joint Conference
  on Artificial Intelligence (2011)

\bibitem{Pellet07}
Sirin, E., Parsia, B., Grau, B.C., Kalyanpur, A., Katz, Y.: Pellet: A practical
  owl-dl reasoner. Web Semant.  5,  51--53 (2007)

\bibitem{ontoQA}
Tartir, S., Arpinar, I.B., Moore, M., Sheth, A.P., Aleman-Meza, B.: {OntoQA}:
  Metric-based ontology quality analysis. In: Proceedings of IEEE Workshop on
  Knowledge Acquisition from Distributed, Autonomous, Semantically
  Heterogeneous Data and Knowledge Sources (2005)

\bibitem{Tsarkov06}
Tsarkov, D., Horrocks, I.: Fact++ description logic reasoner: System
  description. In: Proceedings of the Third International Joint Conference on
  Automated Reasoning. pp. 292--297 (2006)

\bibitem{TsHP07}
Tsarkov, D., Horrocks, I., Patel-Schneider, P.F.: Optimizing terminological
  reasoning for expressive description logics. J.\ of Automated Reasoning
  39(3),  277--316 (2007)

\bibitem{WangP07}
Wang, T.D., Parsia, B.: Ontology performance profiling and model examination:
  First steps. In: Proceedings of The 6th International Semantic Web
  Conference, 2nd Asian Semantic Web Conference, {ISWC} + {ASWC}. pp. 595--608.
  Korea (2007)

\bibitem{Zhang10}
Zhang, H., Li, Y.F., Tan, H.B.K.: Measuring design complexity of semantic web
  ontologies. J. Syst. Softw.  83(5),  803--814 (2010)

\end{thebibliography}
\end{document}